# Formalization of Malagasy conjugation

**Joro Ny Aina RANAIVOARISON[1], Éric LAPORTE[2], Baholisoa Simone RALALAOHERIVONY[1]**

1: University of Antananarivo, Centre Interdisciplinaire de Recherche Appliquée au Malgache, Madagascar
2: Université Paris-Est, Laboratoire d'informatique Gaspard-Monge CNRS UMR 8049, F77454 Marne-la-Vallée, France

**Abstract**

This paper reports the core linguistic work performed to construct a dictionary-based morphological analyser for Malagasy simple verbs. It uses the Unitex platform (Paumier, 2003) and comprised the contruction of an electronic dictionary for Malagasy simple verbs. The data is encoded on the basis of morphological features. The morphological variations of verb stems and their combination with affixes are formalized in finite-state transducers represented by editable graphs. 78 transducers allow Unitex to generate a dictionary of allomorphs of stems. 271 transducers are used by the morphological analyser of Unitex to recognize the stem and the affixes in conjugated verbs. The design of the dictionary and transducers prioritizes readability, so that they can be extended and updated by linguists.

## 1. Introduction

The construction of a dictionary of Malagasy simple verbs is the core of this work. The Malagasy language is an agglutinative language spoken by 20 millions of people. There are written documents in Malagasy: history, literature, newspapers…, but practically no language resources, language processing or technological exploration of the language. This work is a step in this direction through the construction of an electronic dictionary of simple verbs. The method used to build the dictionary is based on the DELA framework (Laporte, 1993): in practice, features of verbs are described, formalized and integrated into dictionary entries. In the section of results, the dictionary of Malagasy simple verbs (DEMA-VS) and the dictionary of morphological variants of roots (DEMA-VSflx) are presented. The dictionary was tested on a corpus of Malagasy journal obtained from Diwersy (2009). The results of this test are developed in the section of evaluation. In the concluding section, we discuss the potential impact of the dictionary on Malagasy language processing.

## 2. Related work

Malagasy language processing is only beginning. No tools for text processing, information retrieval, information extraction or translation are accessible for users yet. Dictionary-based systems such as a morphological analyser (Dalrymple *et al.*, 2006) and a spell-checker (Raboanary *et al*, 2008) have been developed, but they are not widely used and the dictionaries are not available for research. Berlocher *et al.* (2006) propose an efficient, large-coverage morphological analyser for Korean, an agglutinative language. The analyser is based on readable and updatable resources: an electronic dictionary of stems, transducers of suffixes and transducers of generation of allomorphs. Our work on Malagasy verbs follows the same model.

## 3. Methods

The number of Malagasy verb forms is in the hundreds of thousands. Their roots are about 3790 entries (Rabenilaina, 1985). To construct the dictionary of verbs, we took roots of verbs as a point of departure, inserted them into the dictionary and encoded information so that morphological analysers can recognize automatically in a text their conjugated forms. Most verbs occur in text in conjugated form. The recognition of conjugated verbs requires identifying their roots and their affixes. All verb entries receive 2 pieces of information: a stem class and an affix class. We inserted the stem class, the affix class and the conjugation group in the lexical entries to build practically the DEMA-VS. On the one hand, the stem classes give the morphological variants of the roots; on the other hand, the affix classes give the morphemes attached to them. For example, for the verbs *làlo* “pass”, the stem class allows to recognize forms such as *dàlo, dalóv, lalóv* in conjugated forms such as *mandalo* **act-stat.** “pass”[1], *andalovana* **circ.** “the circumstance where or in which (…) pass”, *lalovana* **loc.** “the place where (…) pass”, and the affix class allows to recognize the affixes attached to the root. The recognition of the affixes is done by the transducers in the DELA directory of Unitex and the processing of stems involves the transducers in the Inflect directory. In this section, we report the construction of the conjugation table and the encoding of stem classes and affix classes.

### 3.1. The construction of the conjugation table

Malagasy conjugation is based on voices (Rajaona, 2004). Verb forms vary according to the values of a feature called voice. Elaborating on the basis of Rajaona (1972, 2004), we define 5 values of the voice feature: active-stative (act.-stat.), objective (obj.), locative (loc.), circumstantial (circ.) and agissive-instrumentive (agi.-inst.). These definitions are based on morphology but syntactic features are also used when morphological features are insufficient to tell the voice for a given verb. Each verb is tested for the existence of the 5 voices, and the resulting combinations of voices define conjugation groups. We built a conjugation table divided in 3 parts for the 3 conjugation groups (gc1, gc2 and gc3) and we specified all the information on the paradigmatic forms of each verb and all the morphemes attached to it. This linguistic information is used to encode the stem class and affix class of a verb.

[1] We added graphical accents in the stems above to specify stress. In text, stress is generally not graphically marked.

## 3.2. Encoding stem classes

The stem class encodes the variations of form of a verb root. It allows Unitex to generate the variants of the root, and then to recognize them in a text. The verbs which have the same variations have the same stem class. For example, *dìngana* "tread" and *elanèlana* "intermediate" belong to the same class (3iv) because they behave exactly the same. Both end with *-na*, accept *-ina* (*dingánina* and *elanelánina*) and in contact with the suffixes, both shift the stress and lose the final vowel (*dìngana/dingán* and *elanèlana/elanelán*). The other verbs have been analysed individually in the same way as *dìngana* and *elanèlana*. 3 fields in the code for the stem class specify 3 pieces of information.

- The presence of one of the *-ka*, *-tra*, *-na* endings (which are not suffixes). The code is 0 if the verb does not end with *-ka*, *-tra*, *-na*; 1 if it ends with *-ka*; 2 if it ends with *-tra*; and 3 if it ends with *-na*.
- The possibility for a verb to receive the suffix of objective voice *-ina*: if a verb accepts *-ina*, then it receives the "i" code in the stem class; if not, it receives the "a" code.
- The phenomena they undergo when they are in contact with affixes.

Figure 1 shows the structure of codes for stem classes.

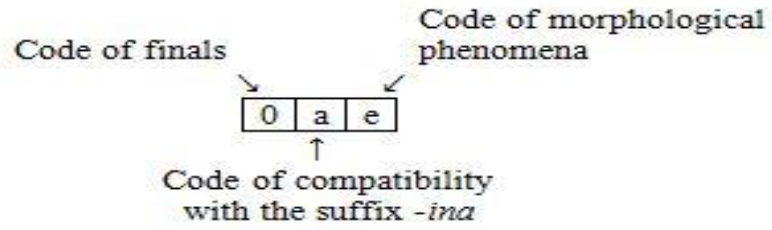

Fig.1: Fields of the code of stem class

Until now, 78 different stem classes have been found and the corresponding 78 inflection transducers (in the sense of Silberztein, 1998) have been built with Unitex. The stem class and the inflection transducer allow Unitex to generate the root forms of a verb, and then to recognize them in conjugated forms in text.

## 3.3. Encoding affix classes

A morphological analyser needs the list of affixes. The conjugation table mentioned in Section 3.1. gives all the affixes and combinations of affixes for each verb. Each verb receives an affix class which specifies which morphemes occur in its conjugated forms. A code with 6 fields is constructed to identify a given affix class. The first field encodes the imperative with *-a*; the second field encodes the active-stative voice and the morpheme of aspect *-aha-*; the third field encodes the agissive-instrumentive voice and the morphemes of aspect *Ø*, *voa-*, *-tafa-*; the fourth field encodes the objective and locative voice; the fifth field encodes the circumstantial voice and the last field encodes the imperative with *-o* and *-y* (Figure 2). When the affixes corresponding to one of the fields do not exist for the verb, the field is filled in with "v". Until now, 271 different affix classes of verbs have been found and the corresponding 271 graphs of combination of morphemes have been constructed to build the morphological analyser. The affix class allows Unitex to recognize affixes in conjugated forms in text.

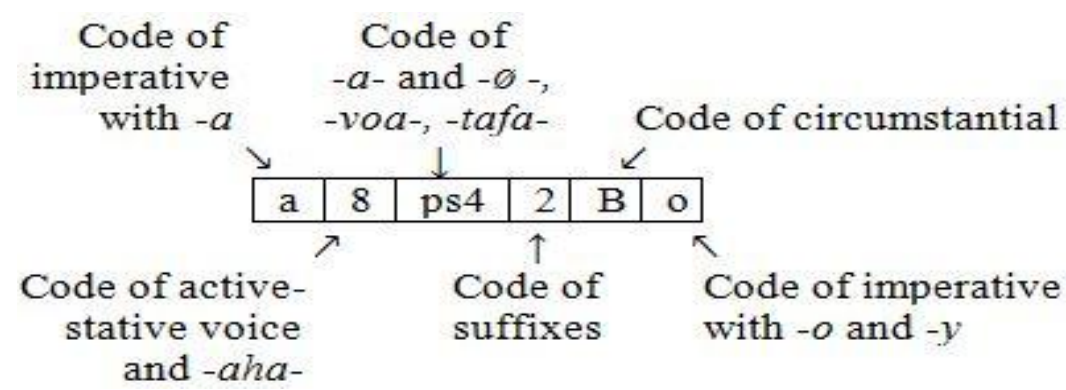

Fig. 2: Fields of the code of affix class

Stem classes and affix classes are lexical information and are the basis of the DEMA-VS dictionary.

# 4. Results

The construction of the stem and affix classes results in the core of an electronic dictionary of Malagasy simple verbs, the DEMA-VS. The combination of a stem class and an affix class provided in the dictionary allows for recognizing at the same time the root and affixes of a verb in a text. Both codes are inserted in the entries of the dictionary with the group of conjugation (gc1, gc2, gc3). The dictionary and graphs are readable, extensible and updatable and are freely available on the web under the LGPLLR license. In this part, we describe the DEMA-VS, the corresponding dictionary of inflected forms (DEMA-VSflx), the transducers and the resources for contractions.

## 4.1. The DEMA-VS dictionary

The DEMA-VS is a morphological and inflectional dictionary of Malagasy simple verbs. At the present time, it contains 536 entries. We selected these entries from conventional dictionaries, so as to cover the most frequent verbs and nearly all existing stem classes and affix classes. The entries of the dictionary provide 3 codes: the stem class, which is preceded by V for 'verb', the affix class, and the conjugation group:

*andràikitra,V2av(1)+a2vvBo+gc1*
*àndrana,V3av(2)+a11ps41Do+gc3*
*andràndra,V0iv+z16ps112Jo+gc3*
*andrìana,V3iv+a16v2Jo+gc3*
*àndro,V0av(1)+a1ps20vAy+gc1*

The algorithms of inflection of Unitex operate directly on the dictionary. They generate the allomorphs of the root and a code that specifies which affixes can be attached to each allomorph.

## 4.2. The DEMA-VSflx dictionary

The DEMA-VSflx is a dictionary of the verb stem forms contained in conjugated verbs. It differs from English or French[2] dictionaries of inflected forms because it does not list complete conjugated forms with their affixes, as *mandro* "take a bath", *androana* "is the circumstance when (…) takes a bath" or *tafandro* "bath completely taken", but

---

[2] The analogous dictionary of inflected forms of simple words for French is the DELAF (Silberztein, 1993, p. 48). It contains nouns, adjectives, verbs, conjunctions, prepositions, etc. For verbs, the inflected forms contained in DELAF are like *aide, aider. V3:P1s:P3s:S1s:S 3s:Y2s*.

gives only the stem of inflected forms and the code that specifies which affixes can be combined with each allomorph of the stem. Below, a sample of entries of DEMA-VSflx is shown:

*andrìana,andrìana.V+a16v2Jo+gc3+0*
*andrián,andrìana.V+a16v2Jo+gc3+ana*
*andrián,andrìana.V+a16v2Jo+gc3+ina*
*andrián,andrìana.V+a16v2Jo+gc3+a*
*andrián,andrìana.V+a16v2Jo+gc3+imprt*
*àndro,àndro.V+a1ps20vAy+gc1+0*
*andró,àndro.V+a1ps20vAy+gc1+ana*
*andró,àndro.V+a1ps20vAy+gc1+a*
*andró,àndro.V+a1ps20vAy+gc1+imprt*

These entries are generated by Unitex from the last 2 entries of the sample of the DEMA-VS given in Section 4.1. The DEMA-VS entry *àndro* "take a bath" produces the last 4 DEMA-VSflx entries above. The combination of these 4 entries with affixes generates 10 conjugated forms:

*androana*
*androy*
*handro*
*handroana*
*hotafandro*
*mandro*
*mandroa*
*nandro*
*nandroana*
*tafandro*

Unitex does not generate any of these conjugated forms, but can recognize them through morphological analysis when they occur in text. Viewing Malagasy as an inflectional language would lead to construct such a dictionary of conjugated forms. At the present time, the DEMA-VSflx contains about 3000 entries, which means that a verb root in Malagasy generates an average of 5.6 morphological variants. Here, with *àndro*, the root generates 4 variants which can be recognized in 10 conjugated forms.

## 4.3. Transducers

To build the DEMA-VS, 2 types of transducers have been constructed: the first for the generation of allomorphs or the variants of the roots and the second for the combination of morphemes. Both types of transducers are presented below in this order.

### 4.3.1. Transducers of generation of stem allomorphs

Verbs in Malagasy can be sorted in 2 types depending on whether they accept the *-ina* suffix or not. For example, the verb *rày* "take" accepts *-ina* and the verb *fàfy* "sow" does not accept it, but only the *-ana* suffix. Accordingly, we constructed 2 types of transducers of generation of allomorphs so that the first type generates the allomorph that combines with *-ina* and the second only the allomorph that combines with *-ana*. The transducer of Fig. 3 (V0are) corresponds to the *0are* stem class and is an example of a transducer of generation of allomorphs for the verbs which do not accept *-ina*. More specifically, this transducer V0are is for those verbs which (i) do not end with *-ka*, *-tra*, *-na*, (ii) do not accept the *-ina* suffix, (iii) lose the first letter in certain conditions specified in the transducer and (iv) have the letter *z* appended in some variants.

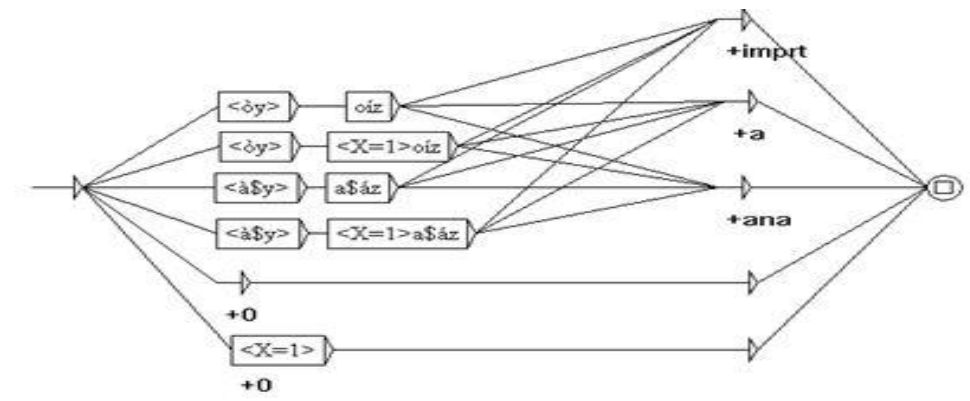

Fig. 3: Example of a transducer of generation of allomorphs

The transducer specifies the information required by Unitex to generate allomorphs of such verbs. Each path of the transducer generates an allomorph. For example, for *fàfy*, V0are generates allomorphs such as *fàfy*, *-áfy*, *-afáz-*, *fafáz-* that can be observed in the conjugated forms *mifafy*, *mamafy*, *mamafaza*, *fafazo*. The morphological information +ana, +imprt, +a, +0 in the right part of the transducer allows to link the allomorphs with the affixes found in the conjugated forms in a text.

### 4.3.2. Transducers of combination of morphemes

In Malagasy, the category which shows most morpho-syntactic variation is verbs. The variations of roots are given by the transducers of generation of allomorphs and the possibilities of combination of these allomorphs with affixes are given by the transducers of morphemes. Grammatical morphemes can be prefixes or suffixes in Malagasy. All kinds of inflectional affixes have been identified (aspect, voice, tense prefixes; imperative, voice suffixes), assigned mnemonic labels and inserted in transducers of combination of morphemes. Such a transducer recognizes all the inflected forms of the verbs that belong to a given affix class. Each path of the transducer recognizes an inflected form by successively checking prefixes, the root (through a node recognizing allomorphs), and suffixes. In Fig. 4, an example is given.

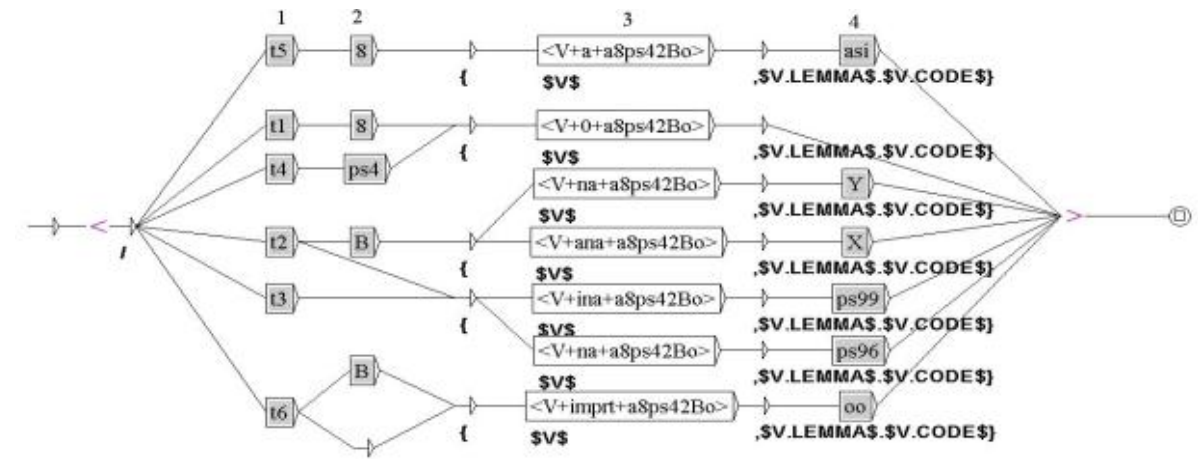

Fig. 4: Example of a transducer of combination of morphemes

The nodes in column 1 give all the prefixes of tense, the nodes in column 2 give all the prefixes of voice and aspect, and the column 3 allow to link variant forms of roots to the suffixes in column 4.

This example corresponds to the *a8ps42Bo* affix class and allows to recognize those verbs which behave as encoded by *a8ps42Bo*: (i) have *-a* and *-o* as a morpheme of imperative, (ii) *-i-* as a prefix of voice, (iii) *-aha-* and *-voa-* as prefixes of aspect and (iv) *-ina* as a suffix of voice. Examples of such verbs are *tàhiry* "preserve", *tèty* "cover", *fidy* "choose", etc. The conjugation of *tàhiry*, for example, gives, in the present tense, forms as *mitahiry*, *mitahiriza*, *mahatahiry*, *voatahiry*, *tehirizina* or *tahirizina*, *tehirizo* or *tahirizo*, etc. *Tèty*, *fidy* behave like *tàhiry*. The transducer allows to recognize also the other two tenses of Malagasy: past and future.

## 4.4. Resources for contractions

In certain conditions, verb forms followed by an invariable word undergo a contraction which involves a graphical elision. For example, *nojereny* is a contraction of *nojerena* “have been watched” and *ny* “by him”. A dictionary of invariable forms (DEMA-INVflx) and transducers for recognition of personal pronouns and elisions have been constructed to increase the recall of the analyser.

### 4.4.1. The DEMA-INVflx

The dictionary of invariable words contains personal pronouns when they are subjects or objects, numerals from one to ten, and a few conjunctions, determiners and demonstrative pronouns. Below, a sample of the DEMA-INVflx is given.

*aho,PRO(NV)+pers:1s*
*ary,CONJC(NV)*
*dimy,DET(NV)+num*
*i,ART(NV)+pers:s*

### 4.4.2. Transducers for verbs with personal pronouns or elisions

In Malagasy, many verb forms are contracted with personal pronouns as in:

(1) *Nojereny aho.*
lit. **Past-** watch-**obj.**-he I.
I have been watched by him.

*Nojereny* analyses as (i) *no-*, a marker of the past tense, (ii) *-jeré-*, a form of the root *jèry* “watch”, (iii) an empty form of *-ana-*, a marker of the objective voice, and (iv) *-ny*, a personal pronoun contracted with the verb. In the absence of *ny*, the form of the *-ana-* marker would be *-na* instead of zero.

The transducer of Fig. 5 recognizes the ending of contractions such as in (1). It is invoked by the transducer of morpheme combinations for *jèry*. The *prop* node invokes a graph of variable personal pronouns[3].

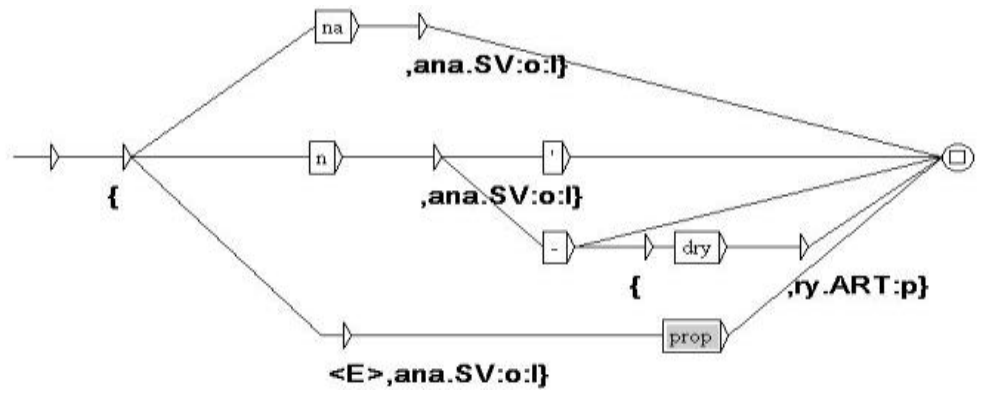

Fig. 5: Example of graph of suffixes recognizing contractions with personal pronouns

In (2), the personal pronoun contracted with the verb in the agissive-instrumentive[4] is recognized by the graph of combination of morphemes displayed in Fig. 6[5].

(2) *Atolony ahy ny tànany*.
lit. **Pres.-**give-**agi.-inst.**-he me **DET** hand-his.
His hand is given to me by him.

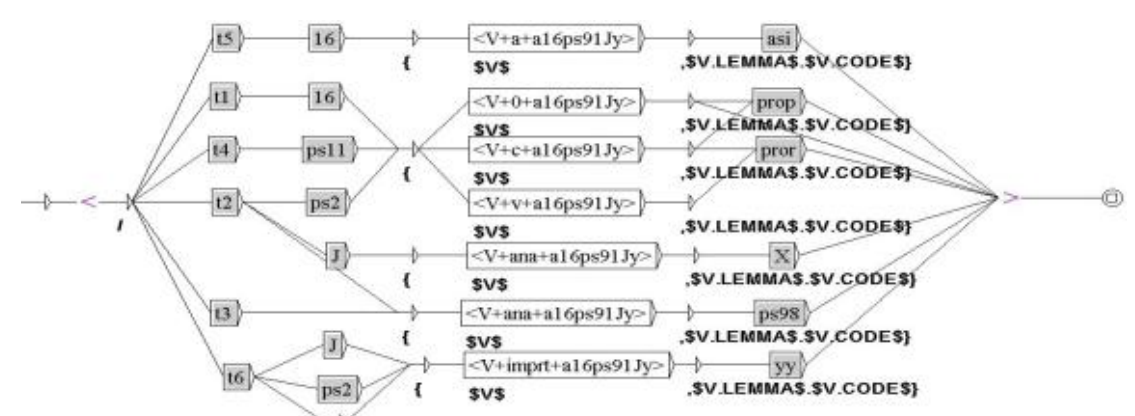

Fig.6: Recognition of personal pronouns with agi.-inst.

Verbs followed by a dash as in *noheverin-* “**past.**-think-**obj.**” or by an apostrophe as in *noraisin’* “**past.**-take **obj.**” are recognized by graphs of suffixes such as that of Fig. 5. Such verb forms occur in sentences such as:

(3) *Noraisin’ny olona ny tànany.*
lit. **past.**-to-take-**obj.**’ **DET** people **DET** hand-his.
His hand is taken by the people.

## 5. Discussion

In inflectional languages, a word cannot generally be parsed into a structure that mirrors its grammatical analysis. For example, the English verb form *masters* is analysed as *master-s*, in which *master* is the stem and *-s* marks at the same time the present and the 3rd person singular. These 3 features: present, 3rd person and singular do not have separate markers. In Malagasy, words can generally be parsed into a structure with separate affixes for each grammatical feature. The verb *mitondra* “wear” is analysed as *m-i-tondra*: *m-* marks the present tense, *-i-* the active-stative voice and *tondra* is the root. This is a major difference between inflectional and agglutinative languages.

However, the number of inflected forms of a given verb in Malagasy, disregarding verb-pronoun contractions, is about 150, which is much less than in Korean[6] and comparable to the number of inflected forms of an adjective in Serbian (Krstev, 2008), a highly inflected language. Therefore, it is certainly possible to generate a dictionary of inflected forms of Malagasy verbs, in the same way as in French, English or Serbian, by combining stem forms and affix-related information. Instead of being viewed as a sequence of morphemes, verbs would then be represented with global values for inflectional features: tense, voice, aspect, mode.

## 6. Evaluation

To evaluate the coverage of the DEMA-VS, the dictionary has been tested on a paragraph of a journalistic corpus of

---

[3] The invariable personal pronouns are inserted in the DEMA-INVflx introduced in 4.4.1.

[4] Both (1) and (2) are in the passive diathesis, but the passive diathesis is marked by the objective voice in (1), whereas it is marked by the agissive-instrumentive voice in (2). The objective voice always marks the passive diathesis, but the agissive-instrumentive voice may mark another diathesis in other verbs.

[5] The *prop* and *pror* nodes in the graph invoke graphs of personal pronouns. 18% of the affix graphs for verbs recognize personal pronouns as shown in Fig. 6 and the other 82% invoke 6 graphs such that of Fig. 5 or DEMA-INVflx.

[6] According to Berlocher *et al.*’s (2006) data, a verb has approximately 120000 inflected forms.

52 000 words (Diwersy, 2009) distributed with Unitex. The paragraph contains 16 sentences and 43 verb occurrences. It belongs to a part of the corpus (cjmc3) that had not been used to construct the dictionary. The occurrences of the verbs were located manually and sorted into 2 classes: 28 (65%) have entries in the dictionary, and 15 have not.

**1. Entry in DEMA-VS**

1. *Hanao*
2. *niakatra*
3. *voampanga*
4. *nanao* *2
5. *nahafahany*
6. *nivoaka*
7. *voalazan'*
8. *nahazoany*
9. *nahatratra*
10. *marihina*
11. *tratry*
12. *nampodiana*
13. *niaiky*
14. *nifona*
15. *hivoaka*
16. *nampiasa*
17. *nanamarika*
18. *ibabohana*
19. *mody*
20. *nentin'*
21. *nataony*
22. *nilaza*
23. *hanomezana*
24. *manaotao*
25. *hanaovan-*
26. *mijanona*
27. *manao*

**2. No entry in DEMA-VS**

1. *voaheloka* *2
2. *nanosika* *2
3. *nampiharin'*
4. *novakiana*
5. *tototry*
6. *nitsarana*
7. *nisalorany*
8. *nisy*
9. *nolaviny*
10. *notsoriny*
11. *nangataka*
12. *mihantona*
13. *nanipy*

### 6.1. Occurrences which have entries in the dictionary

Among the 28 occurrences which have entries in the DEMA-VS, 25 (89%) are correctly analysed by Unitex and 3 (11%) are not, even though the entry exists in the dictionary.

These occurrences are:

1. *voalazan'*
2. *tratry*
3. *manaotao*

For #1, the problem is the recognition of an elided form in presence of an affix of aspect. For #2, it is the recognition of a variant of the stem when the word after the verb is the determiner *ny*. For #3, it is the recognition of a reduplicated form. The solution of these problems involves updating graphs.

### 6.2. Occurrences which have no entries in the dictionary

Among the 15 occurrences which do not have any entries in DEMA-VS,

- 4 (27%) belong both to stem classes and affix classes which have already been found and assigned codes,
- 4 (27%) belong to already encoded affix classes,
- 10 (67%) belong to already encoded stem classes (this subset includes the preceding two),
- 5 (33%) belong to new stem classes and new affix classes.

### 6.3. Results of the evaluation

On the sample text, the lexical coverage of the DEMA-VS is estimated at 65%.

The success rate of the morphological analyser is estimated at 58%. This rate takes into account both dictionary-related and graph-related errors.

The coverage of the language resources in terms of stem classes is estimated at 88%, and their coverage in terms of affix classes is estimated at 74%. We found these results encouraging, since the entire system has been built so that language resources are readable and can be extended and updated.

## 7. Conclusion

The other main differences between Malagasy and English or French lie in the characteristics of conjugation, and in sentence order. Firstly, Malagasy sentence order is VSO (Dalrymple *et al.*, 2006), whereas French and English are SVO; secondly, French and English conjugation involves the feature of person whereas Malagasy does not. Person does not affect the structure of verbs in Malagasy or interfere with conjugation. Our dictionary gives access to language processing applications with a good recall and precision. For example, it might lead to the development of an efficient search function for documents in Malagasy in an information retrieval system.

The DEMA-VS is not complete. Its construction must be continued to reach a good lexical coverage and obtain a successful morphological analyser. The work reported in this paper deals only with verbs, but other parts of speech as nouns and adjectives show less morpho-syntactic variability.